\documentclass[10pt,twocolumn,letterpaper]{article}

\usepackage{cvpr}
\usepackage{times}
\usepackage{epsfig}
\usepackage{graphicx}
\usepackage{amsmath}
\usepackage{amssymb}
\usepackage{multirow}
\usepackage{amsmath}
\usepackage[ruled]{algorithm2e}
\usepackage{pifont}% http://ctan.org/pkg/pifont
\usepackage{enumitem}

\DeclareMathOperator*{\argmin}{arg\,min}
\usepackage[pagebackref=true,breaklinks=true,letterpaper=true,colorlinks,bookmarks=false]{hyperref}

\makeatletter\renewcommand\paragraph{\@startsection{paragraph}{4}{\z@}
  {.5em \@plus1ex \@minus.2ex}{-.5em}{\normalfont\normalsize\bfseries}}\makeatother

\newlength\savewidth\newcommand\shline{\noalign{\global\savewidth\arrayrulewidth
  \global\arrayrulewidth 1pt}\hline\noalign{\global\arrayrulewidth\savewidth}}
\cvprfinalcopy % *** Uncomment this line for the final submission

\def\cvprPaperID{5972} % *** Enter the CVPR Paper ID here

\newcommand{\app}{\raise.17ex\hbox{$\scriptstyle\sim$}}

\ifcvprfinal\fi
\begin{document}
%%%%%%%%% TITLE
\title{Adversarial Examples Improve Image Recognition}

\author{
Cihang Xie\textsuperscript{1,2}\footnotemark \quad
Mingxing Tan\textsuperscript{1} \quad
Boqing Gong\textsuperscript{1} \quad
Jiang Wang\textsuperscript{1} \quad
Alan Yuille\textsuperscript{2} \quad
Quoc V. Le\textsuperscript{1} \vspace{.3em}\\
\textsuperscript{1}Google \qquad\qquad \textsuperscript{2}Johns Hopkins University
\vspace{-.25em}
}

\maketitle
 \renewcommand*{\thefootnote}{\fnsymbol{footnote}}
 \setcounter{footnote}{1}
 \footnotetext{Work done during an internship at Google.}
 \renewcommand*{\thefootnote}{\arabic{footnote}}
 \setcounter{footnote}{0}

\maketitle
\thispagestyle{empty}

%%%%%%%%% ABSTRACT
\begin{abstract}
Adversarial examples are commonly viewed as a threat to ConvNets. Here we present an opposite perspective: adversarial examples can be used to \textbf{improve image recognition models} if harnessed in the right manner. We propose AdvProp, an enhanced adversarial training scheme which treats adversarial examples as additional examples, to prevent overfitting. Key to our method is the usage of a separate auxiliary batch norm for adversarial examples, as they have different underlying distributions to normal examples.

We show that AdvProp  improves a wide range of models on various image recognition tasks and performs better when the models are bigger. For instance, by applying AdvProp to the latest EfficientNet-B7 \cite{Tan2019} on ImageNet, we achieve significant improvements on ImageNet (+0.7\%), ImageNet-C (+6.5\%), ImageNet-A (+7.0\%) and Stylized-ImageNet (+4.8\%). With an enhanced EfficientNet-B8, our method achieves the state-of-the-art \textbf{85.5\%} ImageNet top-1 accuracy \textbf{without} extra data. This result even surpasses the best model in \cite{Mahajan2018} which is trained with 3.5B Instagram images (\app 3000$\times$ more than ImageNet) and \app 9.4$\times$ more parameters. Models are available at \url{https://github.com/tensorflow/tpu/tree/master/models/official/efficientnet}.
\end{abstract}

%%%%%%%%% BODY TEXT
\section{Introduction}
Adversarial examples crafted by adding imperceptible perturbations to images, can lead Convolutional Neural Networks (ConvNets) to make wrong predictions. The existence of adversarial examples not only reveals the limited generalization ability of ConvNets, but also poses security threats on the real-world deployment of these models. Since the first discovery of the vulnerability of ConvNets to adversarial attacks \cite{Szegedy2014}, many efforts \cite{cheng2020cat,Goodfellow2015,jin2020manifold,Kannan2018,Kurakin2017,Madry2018,pang2020boosting,stutz2019confidence,Tramer2018,Xie2019,Zhang2019a} have been made to improve network robustness.

%%%%%%%%%%%%%%%%%%%%%%%%%%%%%%%%%%%%%%%%%%%%%%%%%
\begin{figure}[t!]
\centering
\includegraphics[width=\linewidth]{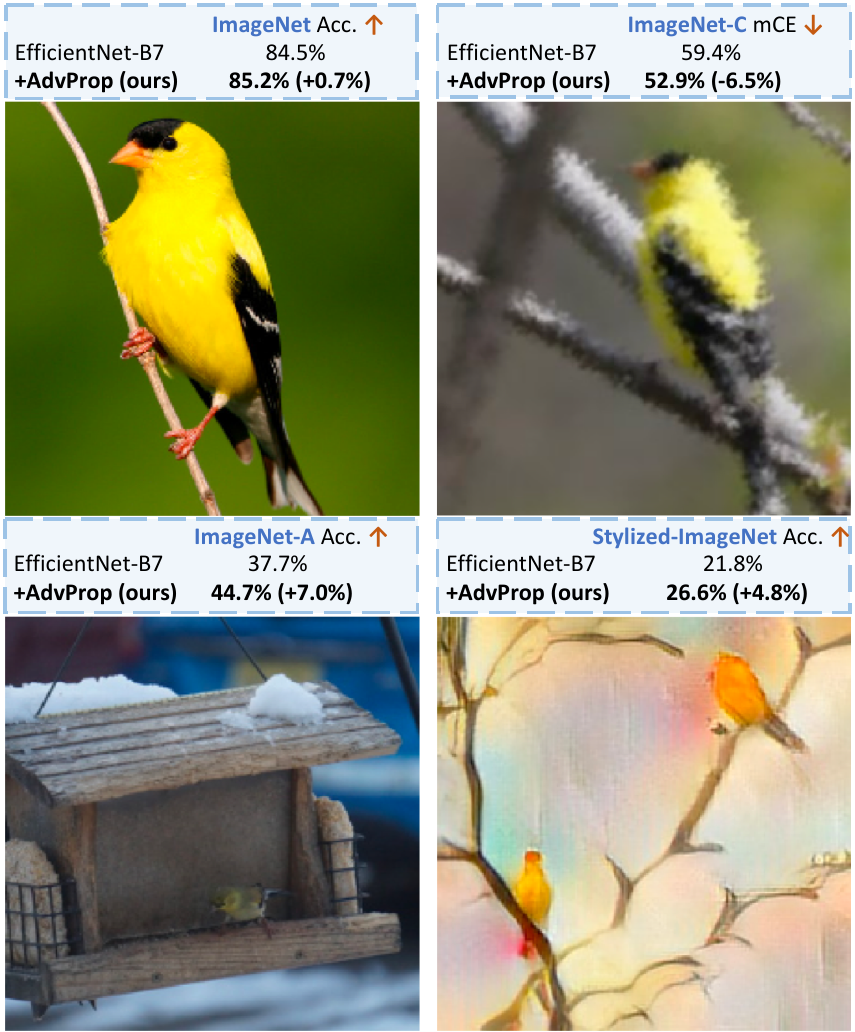}
\caption{\textbf{AdvProp improves image recognition}. By training models on ImageNet, AdvProp helps EfficientNet-B7 \cite{Tan2019} to achieve 85.2\% accuracy on ImageNet \cite{Russakovsky2015}, 52.9\% mCE (mean corruption error, lower is better) on ImageNet-C \cite{Hendrycks2018}, 44.7\% accuracy on ImageNet-A \cite{Hendrycks2019b} and 26.6\% accuracy on Stylized-ImageNet \cite{Geirhos2018}, beating its vanilla counterpart by 0.7\%, 6.5\%, 7.0\% and 4.8\%, respectively. Theses sample images are randomly selected from the category ``goldfinch''.}
\label{fig:teaser}
\vspace{-1em}
\end{figure}
%%%%%%%%%%%%%%%%%%%%%%%%%%%%%%%%%%%%%%%%%%%%%%%%%

In this paper, rather than focusing on defending against adversarial examples, we shift our attention to leveraging adversarial examples to improve accuracy. Previous works show that training with adversarial examples can enhance model generalization but are restricted to certain situations---the improvement is only observed either on small datasets (\eg, MNIST) in the fully-supervised setting \cite{Goodfellow2015,li2019inductive}, or on larger datasets but in the semi-supervised setting \cite{Miyato2018,Qiao2018}. Meanwhile, recent works \cite{Kurakin2017,Kannan2018,Xie2019} also suggest that training with adversarial examples on large datasets, \eg, ImageNet \cite{Russakovsky2015}, with supervised learning  results in performance degradation on clean images. To summarize, it remains an open question of how adversarial examples can be  used effectively to help vision models.

We observe  all previous methods jointly train over clean images and adversarial examples without distinction even though they should be drawn from different underlying distributions. We hypothesize  this distribution mismatch between clean examples and adversarial examples is a key factor that causes the performance degradation in previous works \cite{Kannan2018,Kurakin2017,Xie2019}.

In this paper, we propose AdvProp, short for Adversarial Propagation, a new training scheme that bridges the distribution mismatch with a simple yet highly effective two-batchnorm approach. Specifically, we propose to use two batch norm statistics, one for clean images and one auxiliary for adversarial examples. The two batchnorms properly disentangle the two distributions at normalization layers for accurate statistics estimation. We show this distribution disentangling is crucial, enabling us to successfully improve, rather than degrade, model performance with adversarial examples.

\emph{To our best knowledge, our work is the first to show adversarial examples can improve model performance in the fully-supervised setting on the large-scale ImageNet dataset}. For example, an EfficientNet-B7 \cite{Tan2019} trained with AdvProp achieves 85.2\% top-1 accuracy, beating its vanilla counterpart by 0.8\%. The improvement by AdvProp is more notable when testing models on distorted images. As shown in Fig.~\ref{fig:teaser}, AdvProp helps EfficientNet-B7 to gain an absolute improvement of 9.0\%, 7.0\% and 5.0\% on ImageNet-C \cite{Hendrycks2018}, ImageNet-A \cite{Hendrycks2019b} and Stylized-ImageNet \cite{Geirhos2018}, respectively.

As AdvProp effectively prevents overfitting and performs better with larger networks, we develop a larger network, named EfficientNet-B8, by following similar compound scaling rules in \cite{Tan2019}. With our proposed AdvProp, EfficientNet-B8 achieves the state-of-the-art \textbf{85.5\%} top-1 accuracy on ImageNet \textbf{without} any extra data. This result even surpasses the best model reported in \cite{Mahajan2018}, which is pretrained on 3.5B extra Instagram images (\app 3000$\times$ more than ImageNet) and requires \app 9.4$\times$ more parameters than our EfficientNet-B8.

\section{Related Work}
\paragraph{Adversarial Training.} Adversarial training, which trains networks with adversarial examples, constitutes the current foundation of state-of-the-arts for defending against adversarial attacks \cite{Goodfellow2015,Kurakin2017,Madry2018,Xie2019}. Although adversarial training significantly improves model robustness, how to improve clean image accuracy with adversarial training is still under-explored. VAT \cite{Miyato2018} and deep co-training \cite{Qiao2018} attempt to utilize adversarial examples in semi-supervised settings, but they require enormous extra unlabeled images. Under supervised learning settings, adversarial training is typically considered hurting accuracy on clean images \cite{raghunathan2019adversarial}, \eg, \app10\% drop on CIFAR-10 \cite{Madry2018} and \app15\% drop on ImageNet \cite{Xie2019}. Tsipras \etal \cite{Tsipras2018} argue that the performance tradeoff between adversarial robustness and standard accuracy is provably inevitable, and attribute this phenomenon as a consequence of robust classifiers learning fundamentally different feature representations than standard classifiers. Other works try to explain this tradeoff phenomenon from the perspective of the increased sample complexity of adversary \cite{stutz2019disentangling,min2020curious,nakkiran2019adversarial}, the limited amount of training data \cite{carmon2019unlabeled,najafi2019robustness,schmidt2018adversarially,alayrac2019labels,zhai2019adversarially}, or network overparameterization \cite{raghunathan2020understanding}.

This paper focuses on standard supervised learning without extra data. Although using similar adversarial training techniques, we stand on an opposite perspective to previous works---we aim at using adversarial examples to improve clean image recognition accuracy.

\paragraph{Benefits of Learning Adversarial Features.} Many works corroborate that training with adversarial examples brings additional features to ConvNets. For example, compared with clean images, adversarial examples make network representations align better with salient data characteristics and human perception \cite{Tsipras2018}. Moreover, such trained models are much more robust to high frequency noise \cite{Yin2019}. Zhang \etal \cite{Zhang2019c} further suggest these adversarially learned feature representations are less sensitive to texture distortions and focus more on shape information.

Our proposed AdvProp can be characterized as a training paradigm which fully exploits the complementarity between clean images and their corresponding adversarial examples. The results further suggest that adversarial features are indeed beneficial for recognition models, which agree  with the conclusions drawn from these aforementioned studies.

\paragraph{Data augmentation.} Data augmentation, which applies a set of label-preserving transformations to images, serves as an important and effective role to prevent networks from overfitting \cite{Krizhevsky2012,Simonyan2015,He2016}. Besides traditional methods like horizontal flipping and random cropping, different augmentation techniques have been proposed, \eg, applying masking out \cite{Devries2017} or adding Gaussian noise \cite{lopes2019improving} to regions in images, or mixing up pairs of images and their labels in a convex manner \cite{Zhang2017a}. Recent works also demonstrate that it is possible to learn data augmentation policies automatically for achieving better performance on image classification \cite{Cubuk2018,Cubuk2019,Lemley2017,Lim2019,zhang2019adversarial} and object detection \cite{Zoph2019,Cubuk2019}.

Our work can be regarded as one type of data augmentation: creating additional training samples by injecting noise. However, all previous attempts, by augmenting either with random noise (\eg, Tab. 5 in \cite{Kurakin2017} shows the result of training with random normal perturbations) or adversarial noise \cite{Kannan2018,Kurakin2017,Tramer2018}, fail to improve accuracy on clean images.

\section{A Preliminary Way to Boost Performance}
\label{sec:simple_strategy}
\begin{figure}[t!]
\centering
\resizebox{\linewidth}{!}{
\includegraphics{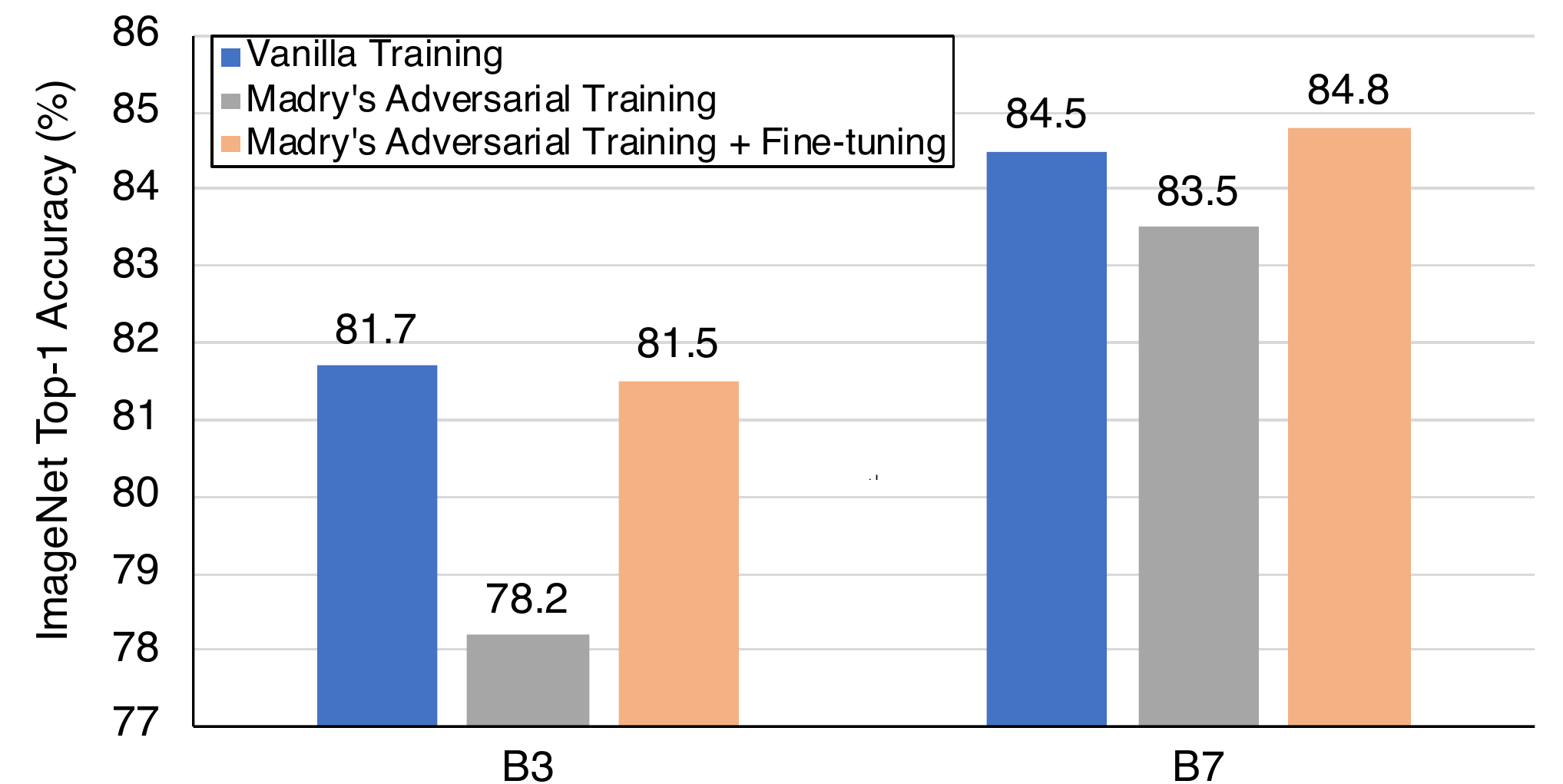}
}
% \vspace{-1.5em}
\caption{Two take-home messages from the experiments on ImageNet: (1) training exclusively on adversarial examples results in performance degradation; and (2) simply training with adversarial examples and clean images in turn can improve network performance on clean images. Fine-tuning details: we train networks with adversarial examples in the first 175 epochs, and then fine-tune with clean images in the rest epochs.}
\label{fig:motivation} 
\vspace{-1.1em}
\end{figure}

Madry \etal \cite{Madry2018} formulate adversarial training as a min-max game and train models exclusively on adversarial examples to effectively boost model robustness. However, such trained models usually cannot generalize well to clean images as shown in \cite{Madry2018,Xie2019}. We validate this result by training a medium-scale model (EfficientNet-B3) and a large-scale model (EfficientNet-B7) on ImageNet using PGD attacker\footnote{For PGD attacker, we set the maximum perturbation per pixel $\epsilon$=4, the step size $\alpha$=1 and the number of attack iteration $n=5$.} \cite{Madry2018}---both adversarially trained models obtain much lower accuracy on clean images compared to their vanilla counterparts. For instance, such adversarially trained EfficientNet-B3 only obtains an accuracy of 78.2\% on the clean images, whereas vanilla trained EfficientNet-B3 achieves 81.7\% (see Fig.~\ref{fig:motivation}).

%%%%%%%%%%%%%%%%%%%%%%%%%%%%%%%%%%%%%%%%%%%%%%%%
%%%%%%%%%%%%%%%%%%%%%%%%%%%%%%%%%%%%%%%%%%%%%%%%

We hypothesize such performance degradation is mainly caused by \emph{distribution mismatch}---adversarial examples and clean images are drawn from two different domains therefore training exclusively on one domain cannot well transfer to the other. If this distribution mismatch can be properly bridged, then performance degradation on clean images should be mitigated even if adversarial examples are used for training. 
To validate our hypothesis, we hereby examine a simple strategy---pre-train networks with adversarial examples first, and then fine-tune with clean images.

The results are summarized in Fig.~\ref{fig:motivation}. As expected, this simple fine-tuning strategy (marked in light orange) always yields much higher accuracy than Madry's adversarial training baseline  (marked in grey), \eg, it increases accuracy by 3.3\% for EfficientNet-B3. Interestingly, while compared to the standard vanilla training setting where only clean images are used (marked in blue), this fine-tuning strategy sometimes even help networks to achieve superior performance, \eg, it increases EfficientNet-B7 accuracy by 0.3\%, achieving 84.8\% top-1 accuracy on ImageNet.

The observation above delivers a promising signal---adversarial examples can be beneficial for model performance if harnessed properly. Nonetheless, we note that this approach fails to improve performance in general, \eg, though such trained EfficientNet-B3 significantly outperforms the Madry's adversarial training baseline, it is still slightly below (-0.2\%) the vanilla training setting. Therefore, a natural question arises: is it possible to distill valuable features from adversarial examples in a more effective manner and boost model performance further generally?

\section{Methodology}
The results in Sec.~\ref{sec:simple_strategy} suggest that properly integrating information from both adversarial examples and clean images even in a simple manner improves model performance. However, such fine-tuning strategy may partially override features learned from adversarial examples, leading to a sub-optimal solution.
To address this issue, we propose a more elegant approach, named AdvProp,
to jointly learn from clean images and adversarial examples. Our method handles the issue of distribution mismatch via explicitly decoupling batch statistics on normalization layers, and thus enabling a better absorption from both adversarial and clean features. In this section, we first revisit the adversarial training regime in Sec.~\ref{Sec:advTrain}, and then introduce how to enable disentangled learning for a mixture of distributions via auxiliary BNs in Sec.~\ref{sec:twoBN}. Finally, we summarize the training and testing pipeline in Sec.~\ref{sec:AdvReg}.

\subsection{Adversarial Training}
\label{Sec:advTrain}
We first recall the vanilla training setting, and the objective function is
\begin{equation}\label{eq:vanilla_training}
\argmin_{\theta}
\mathbb{E}_{(x, y) \sim \mathbb{D}}
\Bigl[ 
L(\theta, x, y)
\Bigr],
\end{equation}
where $\mathbb{D}$ is the underlying data distribution,  $L(\cdot, \cdot, \cdot)$ is the loss function, $\theta$ is the network parameter, and $x$ is training sample with ground-truth label $y$. 

Consider Madry's adversarial training framework \cite{Madry2018}, instead of training with original samples, it trains networks with maliciously perturbed samples, 
\begin{equation}\label{eq:adv_training}
\argmin_{\theta}
\mathbb{E}_{(x, y) \sim \mathbb{D}}
\Bigl[ 
\max_{\epsilon \in \mathbb{S}}
L(\theta, x + \epsilon, y)
\Bigr],
\end{equation}
where $\epsilon$ is a adversarial perturbation, $\mathbb{S}$ is the allowed perturbation range. Though such trained models have several nice properties as described in \cite{Zhang2019c,Yin2019,Tsipras2018}, they cannot generalize well to clean images \cite{Madry2018,Xie2019}.

Unlike Madry's adversarial training, our main goal is to improve network performance on clean images by leveraging the regularization power of adversarial examples. Therefore we treat adversarial images as additional training samples and train networks with a mixture of adversarial examples and clean images, as suggested in \cite{Goodfellow2015,Kurakin2017},  
\begin{equation}\label{eq:mixed_adv_training}
\argmin_{\theta} \bigg[ \mathbb{E}_{(x, y) \sim \mathbb{D}} \Big( L(\theta, x, y) + \max_{\epsilon \in \mathbb{S}} L(\theta, x + \epsilon, y) \Big) \bigg].
\end{equation}
Ideally, such trained models should enjoy the benefits from both adversarial and clean domains. However, as observed in former studies~\cite{Goodfellow2015,Kurakin2017}, directly optimizing Eq.~(\ref{eq:mixed_adv_training}) generally yields lower performance than the vanilla training setting on clean images. We hypothesize that the distribution mismatch between adversarial examples and clean images prevents networks from accurately and effectively distilling valuable features from both domains. Next, we will introduce how to properly disentangle different distributions via our auxiliary batch norm design.

\subsection{Disentangled Learning via An Auxiliary BN}
\label{sec:twoBN}
Batch normalization (BN) \cite{Ioffe2015} serves as an essential component for many state-of-the-art computer vision models \cite{He2016,Huang2016,Szegedy2016a}. Specifically, BN normalizes input features by the mean and variance computed \emph{within each mini-batch}. One intrinsic assumption of utilizing BN is that the input features should come from a single or similar distributions. This normalization behavior could be problematic if the mini-batch contains data from different distributions, therefore resulting in inaccurate statistics estimation.

%%%%%%%%%%%%%%%%%%%%%%%%%%%%%%%%%%%%%%%%%%%%%%%%%
\begin{figure}[h!]
\vspace{-1em}
\centering
\resizebox{0.97\linewidth}{!}{
\includegraphics{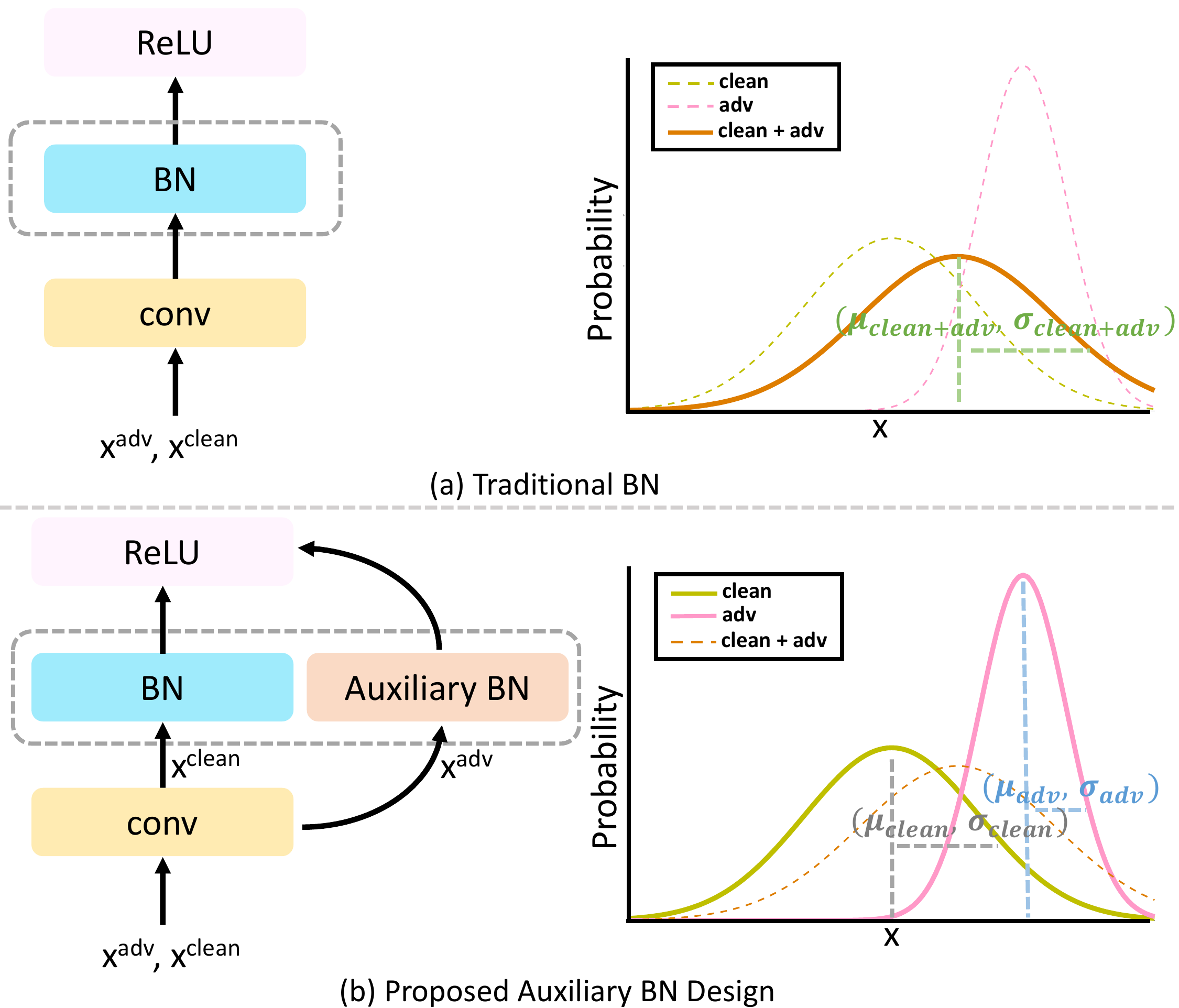}
}
\caption{Comparison between (a) traditional BN usage and (b) the utilization of auxiliary BN. The left and right panels illustrate the information flow in the corresponding network architectures and the estimated normalization statistics when facing a mixture of adversarial and clean images, respectively.}
%\cihang{illustration of auxiliary BN in adversarial training}}
\label{fig:ABN} 
\vspace{-0.35em}
\end{figure}
%%%%%%%%%%%%%%%%%%%%%%%%%%%%%%%%%%%%%%%%%%%%%%%%%

We argue that adversarial examples and clean images have different underlying distributions, and the adversarial training framework in Eq.~\eqref{eq:mixed_adv_training} essentially involves a two-component mixture distribution. 
To  disentangle this mixture distribution into two simpler ones respectively for the clean and adversarial images, we hereby propose an auxiliary BN to guarantee its normalization statistics are exclusively preformed on the adversarial examples. Specifically, as illustrated in Fig.~\ref{fig:ABN}(b), our proposed auxiliary BN helps to disentangle the mixed distributions by keeping separate BNs to features that belong to different domains. Otherwise, as illustrated in Fig.~\ref{fig:ABN}(a), simply maintaining one set of BN statistics results in incorrect statistics estimation, which could possibly lead to performance degradation.

Note that we can generalize this concept to multiple auxiliary BNs, where the number of auxiliary BNs is determined by the number of training sample sources. For example, if training data contains clean images, distorted images and adversarial images, then two auxiliary BNs should be maintained. Ablation studies in Sec.~\ref{sec:attacker_strength} demonstrates that such fine-grained disentangled learning with  multiple BNs can improve performance further.  A more general usage of multiple BNs will be further explored in future works.

\subsection{AdvProp}
\label{sec:AdvReg}
We formally propose AdvProp in Algorithm~\ref{algo:adv} to accurately acquire clean and adversarial features during training. For each clean mini-batch, we first attack the network using the auxiliary BNs to generate its adversarial counterpart; next we feed the clean mini-batch and the adversarial mini-batch to the same network but applied with different BNs for loss calculation, \ie, use the main BNs for the clean mini-batch and use the auxiliary BNs for the adversarial mini-batch; finally we minimize the total loss w.r.t. the network parameter for gradient updates. In other words, except BNs, convolutional and other layers are jointly optimized for both adversarial examples and clean images.

Note the introduction of auxiliary BN in AdvProp only increases a negligible amount of extra parameters for network training, \eg, 0.5\% more parameters than the baseline on EfficientNet-B7. \emph{At test time, these extra auxiliary BNs are all dropped, and we only use the main BNs for inference}.

%%%%%%%%%%%%%%%%%%%%%%%%%%%%%%%%%%%%%%%%%%%%%%%%%
\vspace{-0.9em}
\begin{algorithm}[h!]\small
\DontPrintSemicolon
\KwData{A set of clean images with labels;}
\KwResult{Network parameter $\theta$;\\ }
\For{each training step}{
Sample a clean image mini-batch $x^c$  with label $y$;\\
Generate the corresponding adversarial mini-batch $x^a$ using the auxiliary BNs; \\
Compute loss $L^c(\theta, x^c, y)$ on clean mini-batch $x^c$ using the main BNs;\\
Compute loss $L^a(\theta, x^a, y)$ on adversarial mini-batch $x^a$ using the auxiliary BNs;\\
Minimize the total loss w.r.t. network parameter $\argmin\limits_{\theta} L^a(\theta, x^a, y) + L^c(\theta, x^c, y)$.
}
\KwRet{$\theta$}
\caption{Pseudo code of AdvProp}
\label{algo:adv}
\end{algorithm}
\vspace{-1em}
%%%%%%%%%%%%%%%%%%%%%%%%%%%%%%%%%%%%%%%%%%%%%%%%%

Experiments show that such disentangled learning framework enables networks to get much stronger performance than the adversarial training baseline \cite{Goodfellow2015,Kurakin2017}. Besides, compared to the fine-tuning strategy in Sec.~\ref{sec:simple_strategy}, AdvProp also demonstrates superior performance as it enables networks to jointly learn useful feature from adversarial examples and clean examples at the same time.

\section{Experiments}
\subsection{Experiments Setup}
\paragraph{Architectures.} We choose EfficientNets \cite{Tan2019} at different computation regimes as our default architectures, ranging from the light-weight EfficientNet-B0 to the large EfficientNet-B7. Compared to other ConvNets, EfficientNet achieves much better accuracy and efficiency. We follow the settings in \cite{Tan2019} to train these networks: RMSProp optimizer with decay 0.9 and momentum 0.9; batch norm momentum 0.99; weight decay 1e-5; initial learning rate 0.256 that decays by 0.97 every 2.4 epochs; a fixed AutoAugment policy \cite{Cubuk2018} is applied to augment training images.

\paragraph{Adversarial Attackers.} We train networks with a mixture of adversarial examples and clean images as in Eq.~\eqref{eq:mixed_adv_training}. We choose Projected Gradient Descent (PGD) \cite{Madry2018} under  $L_\infty$ norm as the default attacker for generating adversarial examples on-the-fly. We try PGD attackers with different perturbation size $\epsilon$, ranging from 1 to 4. We set the number iteration for the attackers $n$=$\epsilon$+1, except for the case $\epsilon$=1 where $n$ is set to 1. The attack step size is fixed to $\alpha$=1.

\paragraph{Datasets.} We use the standard ImageNet dataset \cite{Russakovsky2015} to train all models. In addition to reporting performance on the original ImageNet validation set, we go beyond by testing the models on the following  test sets:
\begin{itemize}[leftmargin=*]
    \setlength\itemsep{0.1em}
    \item \textbf{ImageNet-C} \cite{Hendrycks2018}. The ImageNet-C dataset is designed for measuring the network robustness to common image corruptions. It consists of 15 diverse corruption types and each type of corruption has five levels of severity, resulting in 75 distinct corruptions. 
    \item \textbf{ImageNet-A} \cite{Hendrycks2019b}. The ImageNet-A dataset adversarially collects 7,500 natural, unmodified but ``hard'' real-world images. These images are drawn from some challenging scenarios (\eg, occlusion and fog scene)  which are difficult for recognition.
    \item \textbf{Stylized-ImageNet} \cite{Geirhos2018}. The Stylized-ImageNet dataset is created by removing local texture cues while retaining global shape information on natural images via AdaIN style transfer \cite{Huang2017a}. As suggested in \cite{Geirhos2018}, networks are required to learn more shape-based representations to improve accuracy on Stylized-ImageNet. 
\end{itemize}

Compared to ImageNet, images from ImageNet-C, ImageNet-A and Stylized-ImageNet are much more challenging, even for human observers.

\subsection{ImageNet Results and Beyond}
\begin{figure}[t!]
\centering
\resizebox{\linewidth}{!}{
\includegraphics{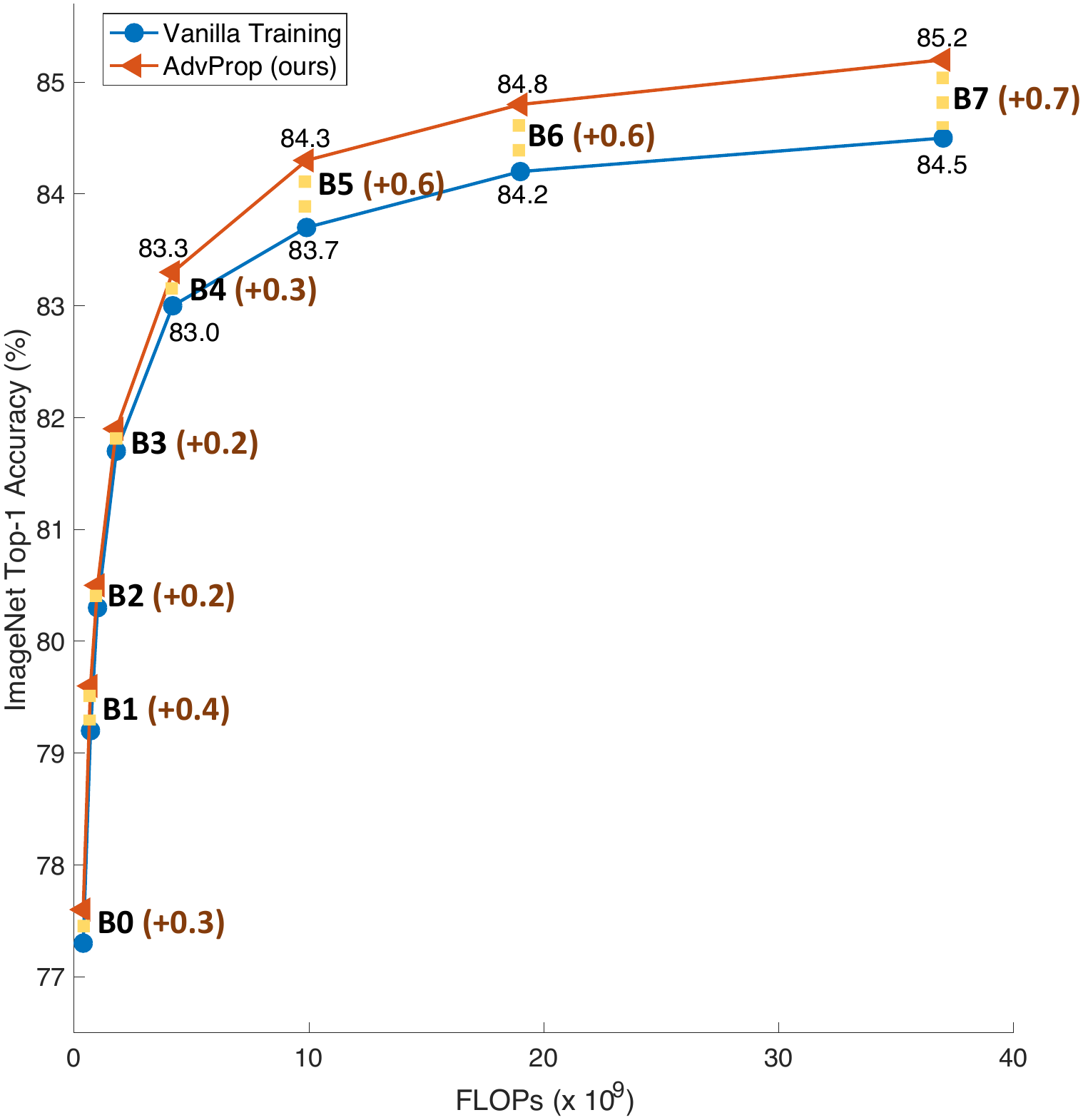}
}
\vspace{-1.4em}
\caption{AdvProp boosts model performance over the vanilla training baseline on ImageNet. 
This improvement becomes more significant if trained with larger networks. Our strongest result is reported by the EfficientNet-B7 trained with AdvProp, \ie, 85.2\% top-1 accuracy on ImageNet.}
\label{fig:overview} 
\vspace{-1.5em}
\end{figure}

\paragraph{ImageNet Results.} 
Fig.~\ref{fig:overview} shows the results on the ImageNet validation set. We compare our method with the vanilla training setting. The family of EfficientNets provides a strong baseline, \eg, EfficientNet-B7's 84.5\% top-1 accuracy is the prior art on ImageNet \cite{Tan2019}.

%%%%%%%%%%%%%%%%%%%%%%%%%%%%%%%%%%%%%%%%%%%%%%%%%
%%%%%%%%%%%%%%%%%%%%%%%%%%%%%%%%%%%%%%%%%%%%%%%%%

As different networks favor different attacker strengths when trained with AdvProp (which we ablate next), we first report the best result in Fig.~\ref{fig:overview}.  Our proposed AdvProp substantially outperforms the vanilla training baseline on all networks. This performance improvement is proportional to the network capacity and larger networks tend to perform better if they are trained with AdvProp. For example, the performance gain is \emph{at most} 0.4\% for networks smaller than EfficientNet-B4, but is \emph{at least} 0.6\% for networks larger than EfficientNet-B4.

Compared to the prior art, \ie, 84.5\% top-1 accuracy, an EfficientNet-B6 trained with AdvProp (with $\app 2\times$ less FLOPs than EfficientNet-B7) already surpasses it by 0.3\%. Our strongest result is obtained by the EfficientNet-B7 trained with AdvProp which achieves 85.2\% top-1 accuracy on ImageNet, beating the prior art by 0.7\%.

%%%%%%%%%%%%%%%%%%%%%%%%%%%%%%%%%%%%%%%%%%%%%%%%
\begin{table}[t!]
\vspace{-0.2em}
\resizebox{1.\linewidth}{!}{
\begin{tabular}{l|c|c|c}
\multirow{2}{*}{Model} & ImageNet-C\textsuperscript{*} \cite{Hendrycks2018}  & ImageNet-A \cite{Hendrycks2019b}  & Stylized-ImageNet\textsuperscript{*} \cite{Geirhos2018}  \\ \cline{2-4} 
                       & mCE \textcolor{red}{$\downarrow$}      & Top-1 Acc.   \textcolor{red}{$\uparrow$}      & Top-1 Acc.  \textcolor{red}{$\uparrow$}      \\ \shline
ResNet-50              &  74.8      & 3.1                & 8.0               \\ \hline
EfficientNet-B0        &  70.7      & 6.7                & 13.1               \\ 
+ AdvProp (\textbf{ours})    &  \textbf{66.2 (-4.5)}& \textbf{7.1 (+0.4)}& \textbf{14.6 (+1.5)}        \\ \hline
EfficientNet-B1        &  65.1      & 9.0                & 15.0              \\ 
+ AdvProp (\textbf{ours})    &  \textbf{60.2 (-4.9)}& \textbf{10.1 (+1.1)}& \textbf{16.7 (+1.7)}    \\ \hline
EfficientNet-B2        &  64.1      & 10.8               & 16.8              \\ 
+ AdvProp (\textbf{ours})    &  \textbf{61.4 (-2.7)}& \textbf{11.8 (+1.0)}& \textbf{17.8 (+1.0)}     \\ \hline
EfficientNet-B3        &  62.9      & 17.9               & 17.8              \\ 
+ AdvProp (\textbf{ours})    &  \textbf{57.8 (-5.1)}& \textbf{18.0 (+0.1)}& \textbf{21.4 (+3.6)}     \\ \hline
EfficientNet-B4        &  60.7      & 26.4               & 20.2              \\ 
+ AdvProp (\textbf{ours})    &  \textbf{58.6 (-2.1)}& \textbf{27.9 (+1.5)}& \textbf{22.5 (+1.7)}     \\ \hline
EfficientNet-B5        & 62.3       & 29.4               & 20.8              \\ 
+ AdvProp (\textbf{ours})    & \textbf{56.2 (-6.1)}    & \textbf{34.4 (+5.0)}& \textbf{24.4 (+3.6)}      \\ \hline
EfficientNet-B6        & 60.6       & 34.5               & 20.9              \\ 
+ AdvProp (\textbf{ours})    & \textbf{53.6 (-7.0)}     & \textbf{40.6 (+6.1)}& \textbf{25.9 (+4.0)}     \\ \hline
EfficientNet-B7        & 59.4       & 37.7               & 21.8              \\ 
+ AdvProp (\textbf{ours})    & \textbf{52.9 (-6.5)}     & \textbf{44.7 (+7.0)}& \textbf{26.6 (+4.8)}     \\
\end{tabular}
}
\vskip 0.05in
\caption{AdvProp significantly boost models' generalization ability on ImageNet-C, ImageNet-A and Stylized-ImageNet. 
The highest result on each dataset is 52.9\%, 44.7\% and 26.6\% respectively, all achieved by the EfficientNet-B7 trained with AdvProp. 
*For ImageNet-C and Stylized-ImageNet, as distortions are specifically designed for images of the size 224$\times$224$\times$3, so we follow the previous setup \cite{Geirhos2018,Hendrycks2018} to \emph{always fix the testing image size at the scale of  224$\times$224$\times$3} for a fair comparison. 
}
\label{tab:imagenet_variants}
\vspace{-1.4em}
\end{table}
%%%%%%%%%%%%%%%%%%%%%%%%%%%%%%%%%%%%%%%%%%%%%%%%

\paragraph{Generalization on Distorted ImageNet Datasets.}
Next, we evaluate models on distorted ImageNet datasets, which are much more difficult than the original ImageNet. For instance, though ResNet-50 demonstrates reasonable performance on ImageNet (76.7\% accuracy), it only achieves 74.8\% mCE (mean corruption error, lower is better) on ImageNet-C, 3.1\% top-1 accuracy on ImageNet-A and 8.0\% top-1 accuracy on Stylized-ImageNet.

The results are summarized in Tab.~\ref{tab:imagenet_variants}. Again, our proposed AdvProp consistently outperforms the vanilla training baseline for all models on all distorted datasets. The improvement here is much more significant than that on the original ImageNet. For example, AdvProp improves EfficientNet-B3 by 0.2\% on ImageNet, and substantially boosts the performance by 5.1\% on ImageNet-C and 3.6\% on Stylized-ImageNet.

The EfficientNet-B7 trained with AdvProp reports the strongest results on these datasets---it obtains 52.9\% mCE on ImageNet-C, 44.7\% top-1 accuracy on ImageNet-A and 26.6\% top-1 accuracy on Stylized-ImageNet. These are the best results so far \emph{if models are not allowed to train with corresponding distortions \cite{Geirhos2018} or extra data \cite{Mahajan2018,Xie2019a}}.

To summarize, the results suggest that AdvProp significantly boosts the  generalization ability by allowing models to learn much richer internal representations than the vanilla training. The richer representations not only provide models with global shape information for better classifying Stylized-ImageNet dataset, but also increase model robustness against common image corruptions.

\paragraph{Ablation on Adversarial Attacker Strength.} 
We now ablate the effects of attacker strength used in AdvProp on network performance. Specifically, the attacker strength here is determined by perturbation size $\epsilon$, where larger perturbation size indicates stronger attacker. We try with different $\epsilon$ ranging from 1 to 4, and report the corresponding accuracy on the ImageNet validation set in Tab.~\ref{tab:attacker_strength}.

%%%%%%%%%%%%%%%%%%%%%%%%%%%%%%%%%%%%%%%%%%%%%%%%
\begin{table}[h!]
\vspace{-0.7em}
\resizebox{1.\linewidth}{!}{
\begin{tabular}{l|c|c|c|c|c|c|c|c}
     & B0   & B1   & B2   & B3   & B4   & B5   & B6   & B7   \\ \shline
PGD5 ($\epsilon$=4) & 77.1 & 79.2 & 80.3 & 81.8 & \textbf{83.3} & \textbf{84.3} & \textbf{84.8} & \textbf{85.2} \\ \hline
PGD4 ($\epsilon$=3) & 77.3 & 79.4 & 80.4 & \textbf{81.9} & \textbf{83.3} &\textbf{84.3}&  84.7    &   85.1   \\ \hline
PGD3 ($\epsilon$=2) & 77.4 & 79.4 & 80.4 & \textbf{81.9} & 83.1 &\textbf{84.3}& 84.7     &   85.0   \\ \hline
PGD1 ($\epsilon$=1) & \textbf{77.6} & \textbf{79.6} & \textbf{80.5} & 81.8 & 83.1 &\textbf{84.3}&  84.6    &   85.0  
\end{tabular}
}
\vskip 0.05in
\caption{ImageNet performance of models trained with AdvProp and different attack strength. In general, smaller networks favor weaker attackers, while larger networks favor stronger attackers.}
\label{tab:attacker_strength}
\end{table}
%%%%%%%%%%%%%%%%%%%%%%%%%%%%%%%%%%%%%%%%%%%%%%%%

With AdvProp, we observe that smaller networks generally favor weaker attackers. For example, the light-weight EfficientNet-B0 achieves the best performance by using 1-step PGD attacker with perturbation size 1 (denoted as PGD1 ($\epsilon$=1)), significantly outperforms the counterpart which trained with 5-step PGD attacker with  perturbation size 4 (denoted as PGD5 ($\epsilon$=4)), \ie, 77.6\% v.s. 77.1\%. 
This phenomenon is possibly due to that small networks are limited by their capacity to effectively distill information from strong adversarial examples, even the mixture distributions are well disentangled via auxiliary BNs.

Meanwhile, networks with enough capacity tend to favor stronger attackers. By increasing attacker strength from PGD1 ($\epsilon$=1) to PGD5 ($\epsilon$=4), AdvProp boosts EfficientNet-B7's accuracy by 0.2\%. This observation motivate our later ablation on keeping increasing attackers strength to fully exploit the potential of large networks.

\subsection{Comparisons to Adversarial Training}
\label{sec:when_adv}
As shown in Fig.~\ref{fig:overview} and Tab.~\ref{tab:imagenet_variants}, AdvProp improves models for better recognition than the vanilla training baseline. These results contradict previous conclusions \cite{Kurakin2017,Tramer2018,Kannan2018} that the performance degradation is always observed if adversarial examples are used for training. We hereby provide a set of ablations for explaining this inconsistency. We choose the PGD5 ($\epsilon$=4) as the default attacker to generate adversarial examples during training.

%%%%%%%%%%%%%%%%%%%%%%%%%%%%%%%%%%%%%%%%%%%%%%%%%
\begin{figure}[h!]
\centering
\vspace{-.5em}
\resizebox{\linewidth}{!}{
\includegraphics{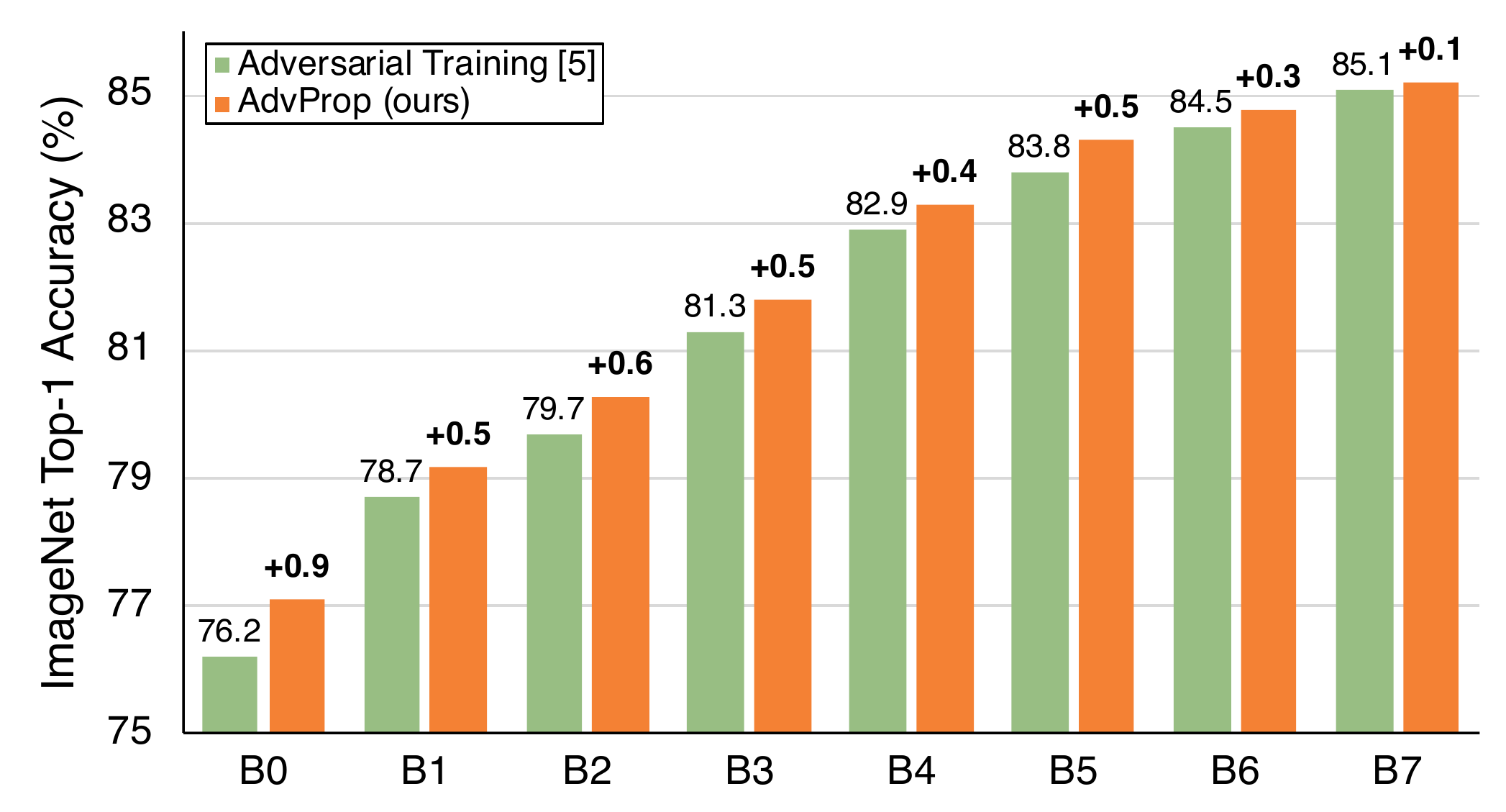}
}
\caption{AdvProp substantially outperforms adversarial training \cite{Goodfellow2015} on ImageNet, especially for small models.}
\label{fig:AT_and_AR} 
\vspace{-1em}
\end{figure}
%%%%%%%%%%%%%%%%%%%%%%%%%%%%%%%%%%%%%%%%%%%%%%%%%

\paragraph{Comparison Results.} 
We compare AdvProp to traditional adversarial training \cite{Goodfellow2015}, and
report evaluation results on ImageNet validation set in Fig.~\ref{fig:AT_and_AR}. Compared to the traditional adversarial training, our method consistently achieves better accuracy on all models. This result suggests that carefully handling BN statistics estimation is important for training better models with adversarial examples.

The biggest improvement is observed when using EfficientNet-B0 where our method beats the traditional adversarial training by 0.9\%. While by using larger models, this improvement becomes smaller---it stays at  \app0.5\% until scaling to EfficientNet-B5, but then drops to 0.3\% for EfficientNet-B6 and 0.1\% for EfficientNet-B7, respectively.

\paragraph{Quantifying Domain Differences.}
One possible hypothesis for the observation above is that more powerful networks have stronger ability to learn a unified internal representations on the mixed distributions, therefore mitigate the issue of distribution mismatch at normalization layers even without the help of auxiliary BNs. To support this hypothesis, we take models trained with AdvProp, and compare the performance difference between the settings that use either the main BNs or the auxiliary BNs. As such resulted networks share all other layers except BNs, the corresponding performance gap empirically
captures the degree of distribution mismatch between adversarial examples and clean images. 
We use ImageNet validation set for evaluation, and summarize the results in Tab.~\ref{tab:bn_acc}.

%%%%%%%%%%%%%%%%%%%%%%%%%%%%%%%%%%%%%%%%%%%%%%%%
\begin{table}[h!]
\centering
\vspace{-0.85em}
\resizebox{\linewidth}{!}{
\begin{tabular}{c|c|c|c|c|c|c|c|c}
                  & B0   & B1   & B2   & B3   & B4   & B5   & B6   & B7   \\ \shline
BN                & 77.1 & 79.2 & 80.3 & 81.8 & 83.3 & 84.3 & 84.8 & 85.2 \\ 
Auxiliary BN      & 73.7 & 75.9 & 77.0 & 78.6 & 80.5 & 82.1 & 82.7 & 83.3 \\ \hline
$\bigtriangleup$  & +3.4 & +3.3 & +3.3 & +3.2 & +2.8 & +2.2 & +2.1 & +1.9
\end{tabular}
}
%\vskip 0.05in
\caption{Performance comparison between settings that use either the main BNs and auxiliary BNs on ImageNet.  This performance difference captures the degree of distribution mismatch between adversarial examples and clean images.}
\label{tab:bn_acc}
\vspace{-0.8em}
\end{table}
%%%%%%%%%%%%%%%%%%%%%%%%%%%%%%%%%%%%%%%%%%%%%%%%

By training with larger networks, we observe this performance difference gets smaller. Such gap  for EfficientNet-B0 is 3.4\%, but then is reduced to 1.9\% for EfficientNet-B7. It suggests that the internal representations of adversarial examples and clean images learned on large networks are much more similar than that learned on small networks. Therefore, with a strong enough network, it is possible to accurately and effectively learn a mixture of distributions even without a careful handling at normalization layers.

\paragraph{Why AdvProp?}
For small networks, our comparison shows that AdvProp substantially outperforms the adversarial training baseline. We attribute this performance improvement mainly to the successful disentangled learning via auxiliary BNs.

For larger networks, though the improvement is relatively small on ImageNet, AdvProp consistently outperforms the adversarial training baseline by a large margin on distorted ImageNet datasets. As shown in Tab.~\ref{tab:AR_advantage},  AdvProp improves EfficientNet-B7 by 3.1\% on ImageNet-C, 4.3\% on ImageNet-A and 1.5\% on Stylized-ImageNet over the adversarial training baseline.

%%%%%%%%%%%%%%%%%%%%%%%%%%%%%%%%%%%%%%%%%%%%%%%%
\begin{table}[h!]
\vspace{-0.9em}
\resizebox{1.\linewidth}{!}{
\begin{tabular}{l|c|c|c}
\multirow{2}{*}{Model} & ImageNet-C \cite{Hendrycks2018}  & ImageNet-A \cite{Hendrycks2019b}  & Stylized-ImageNet \cite{Geirhos2018}  \\ \cline{2-4} 
                       & mCE \textcolor{red}{$\downarrow$}      & Top-1 Acc.   \textcolor{red}{$\uparrow$}      & Top-1 Acc.  \textcolor{red}{$\uparrow$}      \\ \shline
B6 + Adv. Training                      &  55.8     &      37.0           &  24.7 \\
B6 + AdvProp (\textbf{ours})             & \textbf{53.6}       & \textbf{40.6} & \textbf{25.9}      \\  \hline
B7 +    Adv. Training                   & 56.0       & 40.4                   & 25.1 \\
B7 + AdvProp (\textbf{ours})             & \textbf{52.9}       & \textbf{44.7} & \textbf{26.6}      \\ 
\end{tabular}
}
%\vskip 0.05in
\caption{AdvProp demonstrates much stronger generalization ability on distorted ImageNet datasets (\eg, ImageNet-C) than the adversarial training baseline for larger models.}
\label{tab:AR_advantage}
\vspace{-0.7em}
\end{table}
%%%%%%%%%%%%%%%%%%%%%%%%%%%%%%%%%%%%%%%%%%%%%%%%

Moreover, \emph{AdvProp enables large networks to perform better if trained with stronger attackers}. For example, by slightly increasing attacker strength from PGD5 ($\epsilon$=4) to PGD7 ($\epsilon$=6), AdvProp further helps EfficientNet-B7 to achieve \textbf{85.3\%} top-1 accuracy on ImageNet. Conversely, applying such attacker to traditional adversarial training decreases EfficientNet-B7's accuracy to 85.0\%, possibly due to a more severe distribution mismatch between adversarial examples and clean images.

In summary, AdvProp enables networks to enjoy the benefits of adversarial examples even with limited capacity. For networks with enough capacity, compared to adversarial training, AdvProp demonstrates much stronger generalization ability and better at exploiting model capacity for improving performance further.

\paragraph{Missing Pieces in Traditional Adversarial Training.}
In our reproduced adversarial training,  we note it is already better than the vanilla training setting on large networks. For example, our adversarially trained EfficientNet-B7 has 85.1\% top-1 accuracy on ImageNet, which beats the vanilla training baseline by 0.6\%. However, previous works  \cite{Kurakin2017,Kannan2018} show adversarial training \emph{always} degrades performance.

Compared to \cite{Kurakin2017,Kannan2018}, we make two changes in our re-implementation: (1) using stronger networks; and (2) training with weaker attackers. For examples, previous works use networks like Inception or ResNet for training, and set the perturbation size $\epsilon$=16; while we use much stronger EfficientNet for training, and limit the perturbation size to a much smaller value $\epsilon$=4. Intuitively, weaker attackers push the distribution of adversarial examples less away from the distribution of clean images, and larger networks are better at bridging domain differences. Both factors mitigate the issue of distribution mismatch, thus making networks much easier to learn valuable feature from both domains.

\subsection{Ablations}
\label{sec:attacker_strength}
\paragraph{Fine-grained Disentangled Learning via Multiple Auxiliary BNs.}
Following \cite{Tan2019}, our networks are trained with AutoAugment \cite{Cubuk2018} by default, which include operations like rotation and shearing. We hypothesize these operations (slightly) shift the original data distribution and propose to add an extra auxiliary BN to disentangle these augmented data further for fine-grained learning. In total, we keep one main BN for clean images \emph{without} AutoAugment, and two auxiliary BNs for clean images \emph{with} AutoAugment and adversarial examples, respectively.

We try PGD attackers with perturbation size ranging from 1 to 4, and report the best result on ImageNet in Tab.~\ref{tab:fine_grained}. Compared to the default AdvProp, this fine grained strategy further improves performance.  It helps EfficientNet-B0 to achieve
\textbf{77.9\%} accuracy with just 5.3M parameters, which is the state-of-the-art performance
for mobile networks. As a comparison, MobileNetv3 has 5.4M parameters with 75.2\% accuracy~\cite{howard2019searching}.
These results encourage the future investigation on more fine-grained disentangled learning with mixture distributions in general, not just for adversarial training.

%%%%%%%%%%%%%%%%%%%%%%%%%%%%%%%%%%%%%%%%%%%%%%%%%
\begin{table}[h!]\centering
\resizebox{\linewidth}{!}{
\begin{tabular}{l|c|c|c|c|c|c|c|c}
                & B0   & B1   & B2   & B3   & B4   & B5 & B6  & B7  \\ \shline
AdvProp              & 77.6 & 79.6 & 80.5 & 81.9 & 83.3 & 84.3 & \textbf{84.8} & \textbf{85.2} \\ \hline
Fine-Grained AdvProp & \textbf{77.9} & \textbf{79.8} &\textbf{80.7}&\textbf{82.0}& \textbf{83.5} & \textbf{84.4} & \textbf{84.8} & \textbf{85.2}
\end{tabular}
}
\vskip 0.05in
\caption{Fine-grained AdvProp substantially boosts model accuracy on ImageNet, especially for small models. We perform fine-grained disentangled learning by  keeping an additional auxiliary BN for AutoAugment images.}
\label{tab:fine_grained}
\vspace{-0.8em}
\end{table}
%%%%%%%%%%%%%%%%%%%%%%%%%%%%%%%%%%%%%%%%%%%%%%%%%

\paragraph{Comparison to AutoAugment.}
Training with adversarial examples is a form of data augmentation. We choose the standard Inception-style pre-processing \cite{Szegedy2015} as baseline, and compare the benefits of additionally applying AutoAugment or AdvProp.
We train networks with PGD5 ($\epsilon$=4) and evaluate performance on ImageNet.

Results are summarized in Tab.~\ref{tab:data_aug}. For small models, AutoAugment is slightly better than AdvProp although we argue this gap can be addressed by adjusting the attacker strength. For large models, AdvProp significantly outperforms AutoAugment. Training with AutoAugment and AdvProp in combination is better than using AdvProp alone.

%%%%%%%%%%%%%%%%%%%%%%%%%%%%%%%%%%%%%%%%%%%%%%%%
\begin{table}[h!]
\vspace{-0.7em}
\resizebox{\linewidth}{!}{
\begin{tabular}{l|c|c|c|c|c|c|c|c}
            & B0   & B1   & B2   & B3   & B4   & B5   & B6   & B7   \\ \shline
Inception Pre-process \cite{Szegedy2015} & 76.8 & 78.8 & 79.8 & 81.0 & 82.6 & 83.2 & 83.7 & 84.0 \\ \hline
+ AutoAugment \cite{Cubuk2018}    & +0.5 & +0.4 & +0.5 & +0.7 & +0.4 & +0.5 & +0.5 & +0.5 \\
+ AdvProp (\textbf{ours})  & +0.3 & +0.3 & +0.2 & +0.4 & +0.3 & +0.8 & +0.9 & +0.9 \\ \hline
+ \textbf{Both} (\textbf{ours}) & +0.3 & +0.4 & +0.5 & +0.8 & +0.7 & +1.1 & +1.1 & +1.2 
\end{tabular}
}
\vskip 0.05in
\caption{Both AutoAugment and AdvProp improves model performance over the Inception-style pre-processing baseline on ImageNet. Large Models generally perform better with AdvProp than AutoAugment. Training with a combination of both is better than using AdvProp alone on all networks.}
\vspace{-0.7em}
\label{tab:data_aug}
\end{table}
%%%%%%%%%%%%%%%%%%%%%%%%%%%%%%%%%%%%%%%%%%%%%%%%

\paragraph{Attackers Other Than PGD.} 
We hereby study the effects of applying different attackers in AdvProp on model performance. Specifically, we try two different modifications on PGD: (1) we no longer limit the perturbation size to be within the $\epsilon$-ball, and name this attacker to Gradient Descent (GD) as it removes the projection step in PGD; or (2) we skip the random noise initialization step in PGD, turn it to I-FGSM \cite{Kurakin2017}. Other attack hyper-parameters are unchanged: the maximum perturbation size $\epsilon$=4 (if applicable), number of attack iteration n=5 and attack step size $\alpha$=1.0.

For simplicity, we only experiment with EfficientNet-B3, EfficientNet-B5 and EfficientNet-B7, and report the ImageNet performance in Tab.~\ref{tab:other_atatcker}. We observe that all attackers substantially improve model performance over the vanilla training baseline. This result suggests that our AdvProp is not designed for a specific attacker (\eg, PGD), but a general mechanism for improving image recognition models with different adversarial attacker.

\begin{table}[h!]
\vspace{-0.5em}
\centering
\resizebox{0.52\linewidth}{!}{
\begin{tabular}{l|c|c|c}
       & B3   & B5   & B7   \\ \shline
Vanilla Training &  81.7 & 83.7  & 84.5 \\ \hline
PGD \cite{Madry2018}   & 81.8 & 84.3 & 85.2 \\
I-FGSM \cite{Kurakin2017} & 81.9 & 84.3 & 85.2 \\
GD     & 81.7 & 84.3 & 85.3
\end{tabular}
}
\vskip 0.05in
\caption{ImageNet performance when trained with different attackers. With AdvProp, all attackers successfully improve model performance over the vanilla training baseline.}
\label{tab:other_atatcker}
\end{table}

\paragraph{ResNet Results.}
Besides EfficientNets, we also experiment with ResNet~\cite{He2016}. We compare AdvProp against two baselines: vanilla training and adversarial training. We apply PGD5 ($\epsilon$=4) to generate adversarial examples, and follow the settings in \cite{He2016} to train all networks.

We report model performance on ImageNet in Tab.~\ref{tab:resnet}. Compared to vanilla training, adversarial training always degrades model performance while AdvProp consistently leads to better accuracy on all ResNet models. 
Take ResNet-152 for example, 
adversarial training \emph{decreases} the baseline performance by 2.0\%, but our AdvProp further \emph{boosts} the baseline performance by 0.8\%.

%%%%%%%%%%%%%%%%%%%%%%%%%%%%%%%%%%%%%%%%%%%%%%%%
\begin{table}[h!]
\centering
\vspace{-0.82em}
\resizebox{0.95\linewidth}{!}{
\begin{tabular}{l|c|c|c|c}
                     & ResNet-50 & ResNet-101 & ResNet-152 & ResNet-200 \\ \shline
Vanilla Training     &   76.7    &   78.3     &   79.0     &    79.3    \\ \hline
Adversarial Training &   -3.2    &   -1.8     &   -2.0     &    -1.4    \\ 
AdvProp (\textbf{ours})            &\textbf{+0.4}&\textbf{+0.6}&\textbf{+0.8}&\textbf{+0.8}           \\ 
\end{tabular}
}
\vskip 0.02in
\caption{Performance comparison among vanilla training, adversarial training and AdvProp on ImageNet. AdvProp reports the best result on all ResNet models.}
\vspace{-0.55em}
\label{tab:resnet}
\end{table}
%%%%%%%%%%%%%%%%%%%%%%%%%%%%%%%%%%%%%%%%%%%%%%%%

In Sec.~\ref{sec:when_adv}, we show that adversarial training can improve performance if large EfficientNets are used for training. However, this phenomenon is not observed on ResNet, \eg, adversarial training still leads to inferior accuracy even trained with the large ResNet-200. It may suggest that architecture design also plays an important role when training with adversarial example, and we leave it as a future work.

\paragraph{Pushing The Envelope with a Larger Model.} Previous results suggest AdvProp performs better with larger networks. To push the envelope, we train a larger network, EfficientNet-B8, by scaling up EfficientNet-B7 further according to the compound scaling rule in \cite{Tan2019}.

Our AdvProp improves the accuracy of EfficientNet-B8 from 84.8\% to 85.5\%, achieving a new state-of-the-art accuracy on ImageNet without using extra data. This result even surpasses the best model reported in \cite{Mahajan2018}, which is pretrained on 3.5B extra Instagram images (\app 3000$\times$ more than ImageNet) and requires \app 9.4$\times$ more parameters (829M vs. 88M) than our EfficientNet-B8.

\section{Conclusion}
Previous works commonly view adversarial examples as a threat to ConvNets, and suggest training with adversarial examples lead to accuracy drop on clean images.
Here we offer a different perspective: to use adversarial examples for improving accuracy of ConvNets. As adversarial examples have different underlying distributions to normal examples, we propose to use an auxiliary batch norm for disentangled learning by processing adversarial examples and clean images separately at normalization layers. Our method, AdvProp, significantly improves accuracy of all ConvNets in our experiments. Our best model reports the state-of-the-art 85.5\% top-1 accuracy on ImageNet without any extra data.

{\footnotesize
{\noindent {\bf Acknowledgement}: This work was partially supported by ONR N00014-15-1-2356 to Cihang Xie and Alan Yuille.}
}

{\small
\bibliographystyle{ieee_fullname}
\bibliography{egbib}
}
\end{document}